\documentclass{article}

\usepackage{microtype}
\usepackage{graphicx}
\usepackage{subfigure}
\usepackage{booktabs} 

\usepackage{hyperref}

\usepackage[accepted]{icml2024}

\usepackage{amsmath}
\usepackage{amssymb}
\usepackage{mathtools}
\usepackage{amsthm}

\usepackage[capitalize,noabbrev]{cleveref}

\theoremstyle{plain}

\theoremstyle{definition}

\theoremstyle{remark}

\usepackage[textsize=tiny]{todonotes}

\icmltitlerunning{Application-Driven Innovation in Machine Learning}

\begin{document}

\twocolumn[
\icmltitle{Position: Application-Driven Innovation in Machine Learning}



\icmlsetsymbol{equal}{*}

\begin{icmlauthorlist}
\icmlauthor{David Rolnick}{mila}
\icmlauthor{Alan Aspuru-Guzik}{vector}
\icmlauthor{Sara Beery}{mit}
\icmlauthor{Bistra Dilkina}{usc}
\icmlauthor{Priya L.~Donti}{mit}
\icmlauthor{Marzyeh Ghassemi}{mit}
\icmlauthor{Hannah Kerner}{asu}
\icmlauthor{Claire Monteleoni}{inria,cu}
\icmlauthor{Esther Rolf}{harvard,cu}
\icmlauthor{Milind Tambe}{harvard}
\icmlauthor{Adam White}{amii}
\end{icmlauthorlist}

\icmlaffiliation{mila}{McGill University and Mila -- Quebec AI Institute, Montreal, Canada}
\icmlaffiliation{vector}{University of Toronto and Vector Institute, Toronto, Canada}
\icmlaffiliation{mit}{Massachusetts Institute of Technology, Cambridge, USA}
\icmlaffiliation{usc}{University of Southern California, Los Angeles, USA}
\icmlaffiliation{asu}{Arizona State University, Tempe, USA}
\icmlaffiliation{inria}{Inria Paris, Paris, France}
\icmlaffiliation{cu}{University of Colorado Boulder, Boulder, USA}
\icmlaffiliation{harvard}{Harvard University, Cambridge, USA}
\icmlaffiliation{amii}{University of Alberta and Alberta Machine Intelligence Institute, Edmonton, Canada}


\icmlcorrespondingauthor{David Rolnick}{drolnick@cs.mcgill.ca}

\icmlkeywords{applied ML, AI for Good, AI for Science, health, climate}

\vskip 0.3in
]



\printAffiliationsAndNotice{} 

\begin{abstract}
\textbf{In this position paper, we argue that application-driven research has been systemically under-valued in the machine learning community.} As applications of machine learning proliferate, innovative algorithms inspired by specific real-world challenges have become increasingly important. Such work offers the potential for significant impact not merely in domains of application but also in machine learning itself. In this paper, we describe the paradigm of application-driven research in machine learning, contrasting it with the more standard paradigm of methods-driven research. We illustrate the benefits of application-driven machine learning and how this approach can productively synergize with methods-driven work. Despite these benefits, we find that reviewing, hiring, and teaching practices in machine learning often hold back application-driven innovation. We outline how these processes may be improved.
\end{abstract}

\section{Introduction}
Machine learning (ML) is increasingly being used across diverse fields and sectors, with significant impacts for society. ML is being used in healthcare to analyze genetic markers, process medical imagery, and digitize health records \citep{ghassemi2020review}. ML is being used in climate science to speed up physical simulations, parse satellite data, and forecast extreme events \citep{monteleoni2013climate,rolnick2022tackling}. ML is being used in heavy industry to control complex processes, optimize supply chains, and design new materials \citep{gopaluni2020modern}. These kinds of applications, and many others \citep{wang2023scientific}, involve diverse ML tools being used in a multiplicity of ways.

The widespread use of machine learning across society draws on decades of innovation in core ML algorithms, but also on another type of ML research: Application-driven innovation. ML algorithms designed in blue-sky, methods-focused research continue to fall short when used directly for applications. Bridging the gap requires thoughtful consideration of the challenges of real-world tasks and the properties of real-world data, and the welcoming of use cases into the research process. This approach to ML research has much to contribute to broader innovation in ML methods, as well as downstream applications. However, application-driven innovation has characteristics that have led too often to it being under-valued within the ML community or perceived as out-of-scope \citep{rudin2014machine}.

In this paper, we frame the paradigm of application-driven ML (ADML) research, propose where it fits within the ML research landscape, and discuss why it is important not just merely to applications but to advancing ML methods. We reflect on common failures in understanding ADML work during reviewing, hiring, and teaching, and how such factors serve to strongly disincentivize application-oriented work within ML, in both academia and industry research. Finally, we suggest steps toward an ML research ecosystem in which application-driven approaches are recognized and supported alongside methods-driven work.

\section{Paradigms of Innovation in ML}

\subsection{Methods-Driven Research}
The mainstream paradigm of innovation in machine learning methods, which may be referred to as methods-driven research, focuses on the identification of algorithms with certain target properties and evaluates these properties using datasets. Methods-driven ML research has followed various trends over the past several decades (see also \citet{wagstaff2012machine}). For illustration and comparison, we review key aspects of the paradigm as it currently exists, especially in deep learning. 
\begin{figure}
    \centering
    \includegraphics[width=\linewidth]{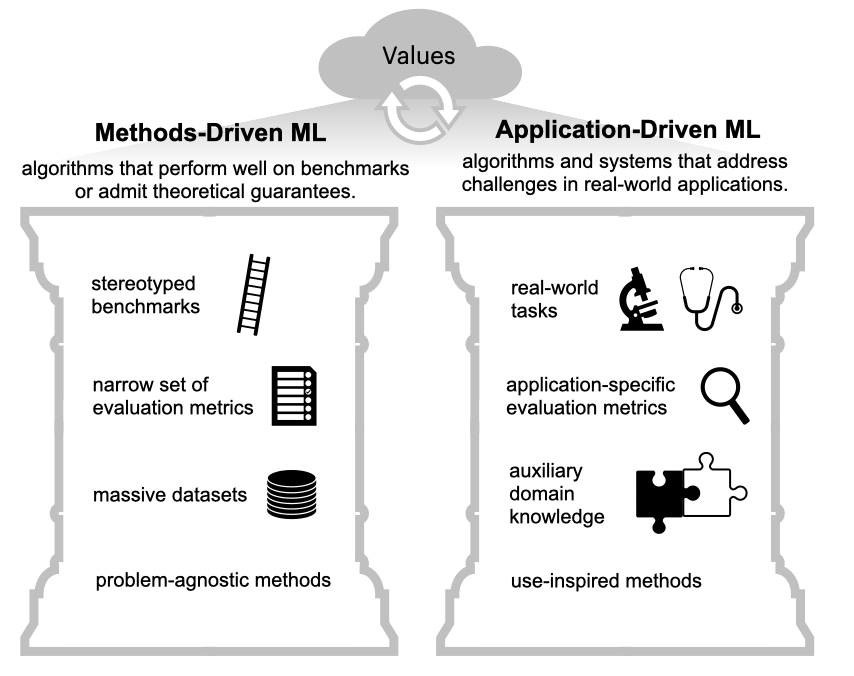}
    \caption{We distinguish between two paradigms of research in machine learning: Methods-driven innovation, in which algorithms are designed based on their performance on standardized benchmarks, and application-driven innovation, in which algorithms are designed to meet challenges faced in real-world problems. We argue that both paradigms contribute significantly to ML research.}
    \label{fig:paradigms}
\end{figure}

\begin{enumerate}
\item \textbf{Stereotyped benchmarks.} Many of the innovations celebrated within the ML community arise through experiments on well-established benchmarks such as ImageNet \citep{deng2009imagenet}, MS COCO \citep{lin2014microsoft}, the OpenAI Gym \citep{brockman2016openai}, etc. While new benchmarks and variations of existing ones continue to be introduced, the number of fundamentally distinct problems remains very limited and the most popular benchmarks have largely retained their hegemony. For example, the three datasets and testbeds above were cited a total of 26,000 times in 2023.
\item \textbf{Narrow set of evaluation metrics.} ML methods are typically evaluated according to a very small set of core metrics, notably test loss and accuracy. Tasks are framed according to a relatively small set of types -- such as fine-grained classification or out-of-distribution generalization -- and within a task type, the method of evaluation generally does not vary or depend on the specific dataset.
\item \textbf{Massive datasets.} Algorithms are assumed to benefit from more data. The subtleties of the data and how it was collected are generally considered to be less important than the quantity \citep{bender2021dangers}, despite mounting evidence that data composition is key \citep{fang2022data}. For example, in creating ImageNet and the training datasets for most large language models, researchers scraped large amounts of raw Internet data, with annotations provided at scale by teams of generally untrained annotators \citep{deng2009imagenet}. Small datasets are generally perceived as less important or trustworthy in assessing an algorithm's effectiveness. 
\item \textbf{Problem-agnostic methods.} As with metrics for evaluation, the algorithms utilized are designed to be as general as possible. It is common to use a variety of very different datasets to evaluate a new method. It is also increasingly common, with the rise in foundation models \citep{bommasani2021opportunities}, to design an algorithm to simultaneously perform a range of different tasks and expect it to generalize well to other tasks.
\end{enumerate}

It is often taken for granted within the ML community that these principles are the only way to produce meaningful innovation in machine learning. Yet as many authors have detailed \citep{birhane2022values,ghassemi2022machine}, these are not inevitabilities but rather are implicit choices that reflect the values and priorities within the field.

\subsection{Application-Driven Research}

Machine learning researchers are frequently shocked to discover that algorithms designed using methods-driven innovation fall short in many real-world use cases. For example, in many remote sensing problems, a simple random forest approach proves more effective than highly regarded computer vision approaches such as Vision Transformers \citep{dosovitskiy2020image}. Such gaps between expectation and reality (which indeed are only apparent when ML researchers engage with end users) illustrate that methods-driven research can be usefully complemented by other types of innovation.

Communities of ML researchers working on problems in remote sensing, health, sustainability, and other areas have developed mature techniques through a paradigm we refer to as \textbf{application-driven machine learning (ADML)}. Below, we detail the core aspects of ADML, as compared to methods-driven research (see Figure \ref{fig:paradigms}).

\begin{enumerate}
\item \textbf{Real-world tasks.} Where methods-driven research evaluates success based on standardized benchmarks, application-driven ML concentrates on a real-world problem, or a family of related problems. The goal is to design an algorithm that satisfies the specific needs of this problem, based on how the ML will ultimately be used. Framing the problem as a machine learning task, or set of tasks, is often part of the challenge. An increasing number of application-specific benchmarks, such as iNaturalist \citep{van2018inaturalist}, ClimSim \citep{yu2023climsim}, WILDS \citep{koh2020wilds}, MIMIC-CXR \citep{johnson2019mimic}, CityLearn \citep{vazquez2020citylearn}, and Grid2Op \citep{marot2021learning}, have been created, often with considerable time and effort. These capture different data and task properties than many standard ML benchmarks and are extremely useful for ADML work. However, no single fixed set of benchmarks is sufficient to capture the full scope of real-world uses for ML.

\item \textbf{Application-specific evaluation metrics.} One of the key reasons that methods-driven ML fails to be impactful in some practical applications is that the set of appropriate evaluation metrics varies greatly between real-world use cases. For example, labeling land cover types from satellite imagery is a common challenge in remote sensing, and while accuracy matters, the computational cost is also a major factor since algorithms must scale over large areas, often on limited computational infrastructure \citep{robinson2019large}. 
Identification of animals from photographs is an important challenge in biodiversity monitoring, and here uncertainty quantification can be essential so that statistics can be appropriately incorporated into ecological models \citep{villon2020new}. From a method-driven ML perspective, both of these tasks are image classification, but for ADML, the downstream criteria for success are different, so evaluation must include these metrics as well as metrics such as test loss and accuracy.

In some cases, standard ML evaluation procedures and frameworks may indeed not apply at all. For example, standard uniform-at-random assignment of instances to test splits can drastically over-estimate performance of a model used in conditions or regions unseen during training \citep{ploton2020spatial,meyer2022machine}; blocked, clustered, or buffered holdout sets are more appropriate to assess performance in out-of-sample use cases for prediction domains with spatial, temporal, or other hierarchical structures \citep{roberts2017cross, le2014spatial}.  It is also worth noting that common goals such as interpretability, robustness, and generalization often have multiple nuanced definitions depending on the particular application. The adversarial robustness metrics commonly used in ML, for example, represent a distinct notion from the Lyapunov stability guarantees expected in certain engineered systems (see e.g.~\citet{donti2022climate}). In applications of reinforcement learning, sample efficiency, sensitivity to hyperparameter choices (because there is no analogy of cross validation), and stability of the deployed controller are especially critical, whereas method-driven evaluation in reinforcement learning often assumes unfettered access to simulator data required to tune ever increasingly complex agents for peak performance. 
\item \textbf{Auxiliary domain knowledge.} While methods-driven ML typically focuses on datasets that are as large as possible, application-driven ML often does not have this luxury. While large datasets remain very useful if they are available, often real-world problems place more data (especially more labeled data) at a premium. In cases where datasets are limited, ADML approaches instead often focus on incorporating as much domain knowledge and auxiliary information as possible, as well as careful selection/curation of labeled data. By contrast, methods-driven research often is unwilling or unable to incorporate such problem-specific information, since distinct domain-specific challenges may be put in a single bucket. For example, if computer vision models pretrained on ImageNet are used on remote sensing data, they must simplify the wide array of sensors used to collect such data, which extend beyond RGB channels. By contrast, ADML approaches for remote sensing explicitly incorporate such diverse inputs \citep{russwurm2024meta,tseng2023lightweight}. Similarly, location information from datapoints is increasingly being incorporated into ADML approaches in areas such as agriculture and ecology \citep{mac2019presence, tseng2022timl,klemmer2023satclip}.

\item \textbf{Problem-informed methods.} Related to the above point on auxiliary data is how the structure of the data and the problem itself is reflected in the choice of algorithm. At the most basic level, many machine learning approaches are chosen based upon a high-level task classification -- thus, convolutional neural networks may be used for image-structured data, LSTMs for time series, etc. However, the structure of a particular problem may be extended beyond such coarse categories. The domain of physics-informed ML often intersects with ADML, if the underlying physics of a problem is considered – e.g., rules such as conservation of energy or mass, or known differential equations governing the variables \citep{kashinath2021physics}. Similarly, ML methods for engineered systems often should incorporate known constraints between variables \citep{donti2021machine}. In cases where ML methods must come with provable guarantees (e.g., in safety-critical systems), incorporating some domain knowledge may in fact be necessary, in order to enable theoretical analysis \citep{donti2020enforcing}. Constraints or rules can often be incorporated into the design of the algorithm, for example, via custom neural network layers or a special loss function \citep{kashinath2021physics}. Desirable objectives can also sometimes be encoded into a loss function, as in the case of optimizing limited resources in public health \citep{wang2023scalable, mate2022field}. Often, input data can be represented in a helpful way with expert knowledge from the application domain, as when climatological variables can be considered as a graph and processed using graph neural networks \citep{cachay2020graph}.
\end{enumerate}
\pagebreak

An important aspect of many ADML innovations is collaboration. To conceive, build, and sustain successful research projects, ADML researchers often need to develop close working relationships and trust with stakeholders relevant to the application, who may be, e.g., researchers in other disciplines, policymakers, or local communities. Significant time is often needed for each project to understand the stakeholder’s needs, translate them to ML opportunities and challenges, and then define research objectives that benefit both the stakeholder and the ML researcher. These steps are essential in framing a real-world problem as an ML task with associated metrics and objectives, in understanding the structure of the data, and in designing algorithms that leverage it.

Note that our discussions of both methods-driven and application-driven research refer to works introducing machine learning methodology, rather than those focused on theory. Of course, many papers include theoretical analysis alongside a largely empirical evaluation, and it is possible for such works to be either methods-driven or application-driven, since theory can be valuable in either case. As an example of theoretical analysis in an ADML work, see \citet{duval2023faenet}, which introduces novel algorithms for materials science applications. Here, the theoretical results justify the property of equivariance under group transformations, which is important for the applications in question.

\subsection{Contributions to Machine Learning}

Paradoxically, while ADML research is problem-centric, this approach has not merely improved the performance of ML in specific use cases, but has done much to advance ML research as a whole. Methods originally tailored for a specific problem have often proven useful to a variety of seemingly dissimilar problems. For example, Fourier Neural Operators \citep{li2020fourier}, originally developed to solve the differential equations governing fluid flow, have now been used in applications as diverse as climate data superresolution and materials property prediction \citep{yang2023fourier,rashid2022learning}. This is counterintuitive, but reflects the fact that real-world challenges may share more underlying structure with each other than with the often artificial benchmarks used in methods-driven research.
\begin{figure}
    \centering
    \includegraphics[width=0.9\linewidth]{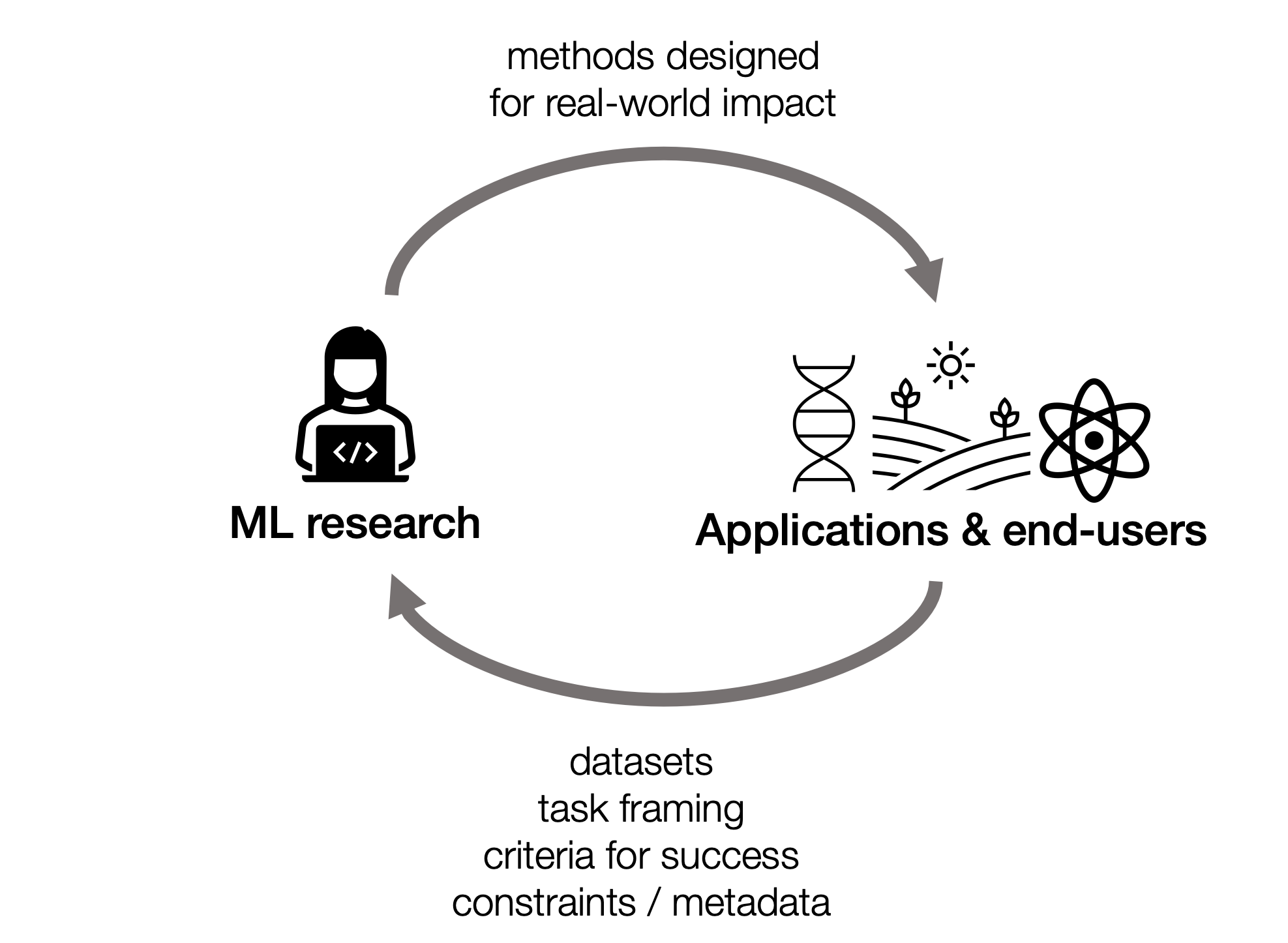}
    \caption{Application-driven innovation in machine learning is achieved through close integration of ML research with applications and end-users. Machine learning methods are designed to be impactful on real-world problems. In turn, applications contribute to ML research methods via novel datasets and task framing, informed by auxiliary domain information such as constraints and metadata. End-users also help define relevant criteria for measuring the success of ML methods on downstream tasks.}
    \label{fig:exchange}
\end{figure}

The use cases raised in ADML can also tie in to broad challenges considered in methods-driven research, or may raise new challenges that can be abstracted away from the task at hand. For example, the iNaturalist benchmark was originally constructed as a byproduct of computer vision research on a real-world biodiversity problem \citep{van2018inaturalist}, but it has now become a popular test case for learning very long-tailed distributions. Many problems in ML for climate and weather modeling pose a special type of nonstationarity challenge, since climate change means that training data drawn from past observations will not necessarily reflect patterns in future data \citep{monteleoni2013climate}.

Finally, ADML has the potential to diversify research directions. Machine learning has often been steered by trends and hype cycles, where many researchers end up working on variations of the same technique or problem. Assessing the relevance of new methods with respect to a wide variety of real-world tasks may be useful, now more than ever, in preventing the community from falling into monolithic patterns of innovation.

\section{Reviewing}
It is currently quite difficult for high-quality ADML work to be published in mainstream machine learning venues. Papers which should fall within the scope of venues such as ICML and NeurIPS are too often excluded, due to systemic weaknesses in the review process. In this section, we explore these weaknesses and suggest how the process may be improved.

Of course, not every paper on applications of ML is well-suited to ICML. Many such papers are most appropriate to publication in the domain of application, or in a cross-disciplinary venue such as Environmental Data Science or Nature Digital Medicine. These papers may, for example, be focused on applications of existing techniques to help address a narrow problem in the domain of interest. However, other papers represent ML contributions -- either in the methodology or in the kind of task being studied. A methodology contribution may be developed for one application but end up being broadly relevant across ML. A task-related contribution may provide a useful testbed for evaluating ML according to real-world challenges, or highlight metrics and considerations not previously regarded as important to ML evaluation.  Both types of contributions thus have much to offer the ML community and fall squarely within the scope of venues such as ICML. Examples of past works of this form which were nevertheless published outside of mainstream ML venues include \citet{ronneberger2015u,raissi2019physics,kurth2023fourcastnet,chen2021enforcing}.

There are a number of common criticisms that will be familiar to anyone who has submitted an ADML paper to a machine learning publication venue:

\begin{itemize}
\item \textbf{Unfamiliar benchmarks.} Reviewers are often confused when a paper does not test against the suite of benchmarks that have become standard in methods-driven ML. Often, ADML papers do not use such benchmarks because the structure of standard datasets, or the goals of standard benchmarking tasks, differ significantly from the real-world data under consideration. For example, image classification algorithms developed for remote sensing are generally not ideal for ImageNet and vice versa. Researchers may also avoid certain standardized benchmarks because of toxic content \citep{prabhu2020large,birhane2021multimodal}, ethics issues involved in their creation \citep{asano2021pass,perrigo2023exclusive}, or known inaccuracies or biases \citep{luccioni2023bugs,crawford2021excavating}. Despite these considerations, reviewers frequently call for experiments on the ``stereotypical'' datasets if they are not familiar with the datasets used in the paper, even when such datasets are high-quality, published, and supported by appropriate baselines.
\item \textbf{Limited applicability.} As noted above, there are certainly applied ML papers with a very narrow focus, and some of these should not be published in ML venues. However, the claim of limited applicability is often directed at application-focused papers indiscriminately, and can indicate unfamiliarity (or hubris) on the part of the reviewer, as when an entire field such as agriculture, ecology, or healthcare is tacitly dismissed as niche. Reviewers also may have unrealistic expectations about the number of benchmarks on which a method is evaluated. For ADML papers, each benchmark may require considerable effort, and there may also be relatively few options for high-quality ML-ready datasets on a given topic, even where the domain problem is very important.
\item \textbf{Too simple.} The common criticism that a novel method is ``too simple'' is often used when the reviewer is surprised that it outperforms complicated ``SoTA'' algorithms. ML scholarship often prizes the appearance of complexity, which can lead to simple insights being disregarded where ideally they would be prized \citep{wagstaff2012machine, lipton2018troubling}. By contrast, ADML researchers often have an especial focus on usability and scalability, so the lack of unnecessary complexity can indeed be a goal, and may be the result of distilling a complex method down to its core innovation. 
\item \textbf{Not innovative.} Application-driven innovation often arises from deep understanding of the challenges posed by a task (which is itself an important contribution) and identifying how to combine or modify existing ingredients in a new way to address this challenge. In such cases, as with ``too simple'', the clear effectiveness of a novel method can be used against it. A surprisingly common review is ``if this method really works so well, then someone must have done it before.''
\end{itemize}

These failure modes of the review process do not necessarily arise from the high-level reviewing guidelines of ML venues, but rather are the consequence of interpreting these guidelines according to the methods-driven research approach detailed above. Below, we attempt to provide guidance to reviewers and area chairs on how to interpret common rubrics for ML research excellence in the case of application-driven work. These may additionally be useful to ML venues interested in specifying reviewing guidelines with a greater degree of granularity. (See also the reviewer guidelines for the AAAI Special Track on AI for Social Impact.)

\begin{itemize}
\item \textbf{Originality.} In the context of ADML research, originality can mean wholly novel methods, a novel combination of existing methods to solve the task at hand, or a novel task (for example where real-world data has a special structure or metrics for success). Note that a new task is not necessarily the same as a new dataset -- it may be a new way of framing or evaluating existing data.
\item \textbf{Quality.} As noted above, ADML methods are generally evaluated on datasets that fall outside the stereotypical benchmarks that many reviewers are familiar with. Such datasets should be sufficient for evaluation if they are well-documented and tested. Algorithms should be compared against classical approaches, which can sometimes be very effective in practice, as well as any applicable ``state-of-the-art'' methods. Ablation experiments for proposed methods are often especially important.
\item \textbf{Clarity.} A common failure of ADML papers is writing for the wrong audience. For publication in an ML venue, a paper should provide background in the relevant application domain and dataset(s), definitions of any jargon used, and clear motivation for the problem at hand. The authors should, of course, also provide ML methodological details to a higher standard than that sometimes expected in applied venues.
\item\textbf{Significance.} This item can be the hardest to agree upon. Different reviewers may have different opinions as to how to weigh the significance to ML and the significance to the domain of application. We would argue that for publication in an ML venue, there must be some importance to the ML community, but significance to another field can be an important factor as well.
\end{itemize}

In addition to following best practices and avoiding common pitfalls, the reviewer pool should include a range of application-informed ML researchers. This can help ensure that application-oriented papers are (i) oriented towards genuinely impactful challenges, (ii) incorporate domain considerations and metrics that are necessary to the use case, and (iii) truly address the problem at hand, rather than merely paying lip-service to impact. Ideally, each ADML paper should have some (but not necessarily all) of its reviewers drawn from ML researchers with experience in the relevant domain of application. Achieving this entails expanding the pool of ADML-aligned researchers invited to review. It must also involve strengthening the reviewer matching process with applications in mind (current keyword-based approaches struggle, e.g.~for terms such as ``forests'' and ``energy'' with both methodological and application-related meanings).

Various ML venues already have strong subcommunities of applied researchers among the author and reviewer pools, but these tend to be limited to certain areas. For example, NeurIPS / ICML include a growing community in ML for materials science, while CVPR / ECCV / ICCV have significant communities in ML for ecology. For these areas of work, the barriers to entry have therefore become somewhat reduced, but are still present, while high barriers remain for many other areas of application.

Overall, we are seeing slow but meaningful progress in ADML publication opportunities. In addition to the growing pool of researchers empowered within the ML community, several ML venues have created focused tracks aimed at specific kinds of ADML publications. The NeurIPS Datasets and Benchmarks track has become a home for papers on innovative application-aligned tasks, while the JMLR special issue on climate change, AAAI Social Impact track, and IAAI conference have similarly broadened the scope of available venues. The further development of special tracks oriented towards application-driven work could be a valuable next step for ML publication venues. We hope, however, that ultimately such workarounds become obsolete as ADML research becomes better recognized and supported as part of mainstream ML innovation.

Many ADML researchers have gravitated towards subcommunities of the broader field of ML, which present opportunities for exchange with like-minded individuals. These spaces include workshop series such as ML4D, ``Tackling Climate Change with Machine Learning'',  and ``RL for Real Life'', and conferences such as ML4H and Climate Informatics. These spaces serve an important purpose complementary to ML venues such as ICML and NeurIPS, furthering discourse between researchers in ML and specific application domains, and in some cases serving as publication venues. However, they are not a replacement for the recognition of ADML research within the wider ML community, and the relegation of such work to smaller, more narrowly focused spaces stands to silo ADML works that would be useful and interesting to ML research at large.

\section{Hiring}
To enable application-driven machine learning, it is not merely sufficient to evaluate ADML research fairly -- it is also necessary to empower ADML researchers. While increasing numbers of trainees in ML are familiar with the ADML paradigm of innovation as well as the traditional methods-driven approach, such individuals continue to face systemic obstacles to advancement in the ML community.
\begin{figure}
    \centering
    \includegraphics[width=\linewidth]{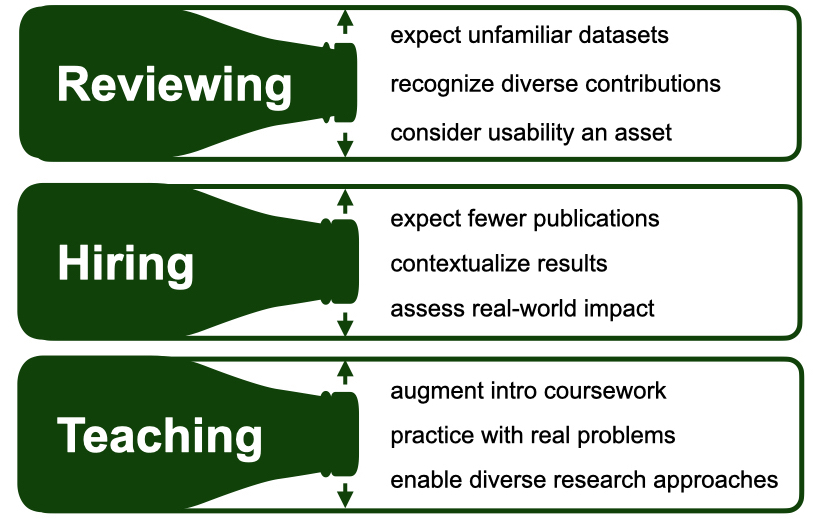}
    \caption{We outline how to alleviate bottlenecks holding back application-driven ML research across reviewing, hiring, and teaching practices. In each of these settings, current practices are largely aligned with methods-driven research and ADML work is often under-valued.}
    \label{fig:bottlenecks}
\end{figure}

Most centers of ML research (in both academia and industry) are overwhelmingly methods-driven. Indeed, many of the senior researchers in this area did not originally pursue this kind of work, but have shifted into more application-driven approaches after gaining their jobs and recognition, or pursue ADML research as merely a part of their toolkit. Generic job opportunities in ML research are generally framed by default along methods-driven principles. While specific opportunities in ML for healthcare, ML for climate, etc. are increasing, these remain a small fraction of the whole, and hiring committees also often do not know how to effectively evaluate ADML candidates.

There are several notable benefits for removing these obstacles and providing equal opportunity to ADML research. At the level of the overall field of machine learning, we have seen that ADML can be an integral part of the ML innovation landscape and synergize with methods-driven approaches. For individual organizations such as computer science departments and industry research labs, ADML can also be a strategic way to increase the impact of research, forge collaborations with other fields, recruit new funders, and demonstrate relevance in the broader space of science and tech. Moreover, the focus of ADML on tasks relevant to a diversity of stakeholders and use cases can often be a boon to institutional goals of equity and inclusion, by reflecting the varied needs of human society and motivating broader engagement in ML innovation.

We now consider the ways in which ADML researchers may not fall into the traditional mold expected within the ML community, with particular emphasis on how research institutions may counteract biases in hiring.

\begin{itemize}
\item \textbf{Publications.} The number of publications for ADML researchers may be lower than is common in methods-driven ML. This is because the innovation pipeline can take longer when extensive work is required to understand, frame, and prepare real-world problems for machine learning. Many ADML projects require creation of new datasets. Collecting and curating datasets in real-world environments can involve extensive time and cost for execution, especially when data collection requires community permission and engagement. Publication venues may also be diverse, not merely the standard ML venues (as we have noted, such venues often present obstacles to publication for ADML work). This need not be a sign that research contributions fall outside of ML -- indeed, many foundational methods such as the U-Net and PINN were first published outside of traditional ML venues \citep{ronneberger2015u,raissi2019physics}.
\item \textbf{Results.} As we have already described, methods developed in ADML may not be tested on standard benchmarks, because these benchmarks may have differently structured data or may not test the relevant algorithmic capabilities. In establishing effectiveness of a result, tests on real-world data are generally more important than theoretical guarantees, which often require unrealistic assumptions.
\item \textbf{Impact.} It is common for ML research to claim useful applications, and indeed many methods-focused papers assume that testing on a variety of standard benchmarks is sufficient to establish far-reaching real-world impacts. However, such assumptions are not always borne out in practice. Application-driven ML is the most likely to be able to justify claims of impact, providing a logical chain deriving the methodology from the ultimate use case. When assessing claims made in application-oriented ML research, it may in some cases be useful to consult researchers in the relevant application domain. A reference letter from an expert outside of computer science may also be helpful. Note that simply counting citations may not adequately capture how impactful a research contribution is in deployed applications.
\end{itemize}

For institutions looking to attract and retain excellent researchers working in application-driven ML, we have several pieces of advice. First, build strong frameworks for interdisciplinary collaboration. This is essential to most ADML work and can serve to enhance the ultimate impact of the research both within ML and in applications. Second, provide support for data engineering teams. Such teams can serve to alleviate time-consuming bottlenecks to the acquisition and restructuring of real-world data to fit machine learning, giving the researchers more time to focus on the algorithms. Finally, strengthen tech transfer pipelines from ML research to startups and other private and public entities. Tech transfer can enable scalable deployments, supporting the overall innovation landscape and raising additional questions for ML research.

\section{Teaching}
The emphasis placed on methods-driven work within the ML community also extends to how the next generation of researchers is trained, leading to a self-perpetuating cycle. Students often come into ML programs especially excited about impactful applications, but paradigms for pedagogy, practice, and research often deprioritize application-driven innovation. We consider the different axes by which this occurs:

\begin{itemize}
\item \textbf{Coursework.} Most machine learning classes present algorithms in isolation. There is very little discussion of how these ML methods will ultimately be used, the characteristics of real-world data, the challenges involved in working with it, and the needs of users and other stakeholders. Performance is often simplified as corresponding to simple numerical measures, and the paradigm of ML innovation is often framed as a progression according to these measures: Method B improves upon the accuracy of Method A, and is in turn superseded by Method C. Such a narrative mischaracterizes innovation as a linear journey along a single path. The phrase ``state-of-the-art'' is often used to characterize algorithms even in the absence of a specific task or dataset, suggesting erroneously that improvements are clear-cut and universal. 
\item \textbf{Hands-on practice.} Training in ML typically involves hands-on projects, both within and outside of coursework. Here the standardized tools that are used often reinforce the methods-driven paradigm by fitting everything into a user-friendly but limited mold. For example, the developer platform Weights \& Biases offers hyperparameter tuning for reinforcement learning algorithms, but these assume that the cost function is deterministic, which is rarely the case in practice. Student-oriented hackathons and Kaggle competitions often frame impact according to leaderboards with simple statistics that abstract away the subtleties of the problem. Such challenges prioritize 
innovations that can be produced in a short timeframe, potentially leading students to assume that impact is something achieved by using the most powerful algorithms on a ready-to-use dataset. Prepackaged challenges implicitly under-emphasize data engineering skills, as the careful data work necessary to tackle real-world problems has already been taken care of by someone else. Working on problems where the data and task are already structured for ML greatly limits the number (and importance) of problems that can be addressed.
\item \textbf{Research.} Within research, students are generally exposed to papers following the methods-driven paradigm. They may learn by experience or example that an application-driven approach is more likely to lead to rejection by major ML publication venues. Understanding a new application domain or unfamiliar research approach takes time. Combined with the pressure to publish frequently, this can make it especially difficult for students to pursue ADML research.
\end{itemize}

With the prevalence of such methods of training, it is small wonder that students predominantly learn methods-driven approaches to research. This represents a loss to the ML community, given the potential for impact and innovation which we have seen that ADML approaches can bring. Moreover, an exclusive focus on methods-driven work can significantly increase attrition among researchers from traditionally underrepresented backgrounds. given that ADML places a clear emphasis on addressing problems and needs of a diverse set of stakeholders, creating an environment that can be both motivating and welcoming. Indeed, ADML-oriented subcommunities such as ``AI for Good'' are often significantly more diverse across gender, ethnicity, and other axes, compared to the ML community as a whole.

Moving forward, the goal for ML education should not be to replace methods-driven approaches, but to augment them. Introductory classes are a particularly important time for students to be exposed to both the methods-driven and application-driven perspectives on machine learning. Courses in specific areas such as AI for Science, Climate, and Health should be encouraged, and some exposure to such areas of work should ideally be a required part of a machine learning degree. In addition, students in computer science or ML should be encouraged to take relevant courses in other disciplines, in order to gain exposure to the different toolkits and ways of thinking within those other fields. Departments should facilitate exchange between students and researchers within and outside of ML, and encourage the development of interdisciplinary collaboration and communication skills. Students should gain at least some practice in the full project lifecycle, taking a problem all the way from framing to deployment. (This is an especially highly valued skillset in industry research, and such experience can provide a significant advantage in competitive industry research labs.) And outside the academic domain, reform in the review system can have a trickle-down effect on the behavior incentivized in students, as can changes in the expectations on the number of publications versus their quality.

It is also worth considering the role of education beyond just preparing the next generation of researchers. Not every student will go into academia or private-sector research. The ML community, to its credit, traditionally recognizes this fact more than many other disciplines (such as Mathematics), with a culture of education that often includes internships, entrepreneurship training, and other preparation for non-research positions. However, such preparation generally focuses on a relatively limited set of roles. The common question posed to students:  ``Academia or industry?'' neglects public sector jobs completely, despite the abundance of public sector positions in, for example, ML for remote sensing, transportation, energy, and weather. Furthermore the ``industry'' that is referred to typically means the tech industry, rather than the ML positions increasingly present in companies outside of tech. These kinds of positions are typically not discussed, advertised, prepared for in classes, or structured as internships, and partnerships that could help train students for such roles are often neglected. Such positions, similar to many industry positions, are tied directly to the deployment of technology and serve as a connection point and a conduit for the translation of impact-driven research across domains. Moving forward, we hope that an application-mindful approach to education broadens the kinds of preparation we are offering our students.

\section{Discussion}

It has been promised that machine learning will be used across society in a myriad ways. Some of the vaunted capabilities for ML are merely techno-optimism and marketing. Some are possible now with present-day methods. And some may be possible in the future. In order for ML research to grow impactfully to meet tomorrow's use cases, we need application-driven innovation.

In this paper, we presented the paradigm of ADML, which complements the traditional paradigm of methods-driven innovation. We outlined and compared the distinguishing features of these two approaches, and why both fall under the broader umbrella of machine learning research. Within ADML, we considered common failure modes in the fair assessment of research quality, and suggested approaches for mitigating these issues. We also discussed bottlenecks to the empowerment of ADML researchers and suggested ways to improve the hiring and retention of these individuals within research positions.

Of course, something we have largely not discussed in this paper is how the ``applications'' driving ML are chosen, nor who assesses what a ``solution'' entails. Every application and ML method represents choices, either explicitly or implicitly made, based on values and priorities. Some use cases that drive ML innovation have detrimental effects on society, and the ways in which ``solutions'' are created may lead to negative or ambiguous outcomes. We largely do not attempt to address these topics in this paper, but point the reader to excellent treatments of related issues in e.g.~\citet{birhane2021algorithmic,o2017weapons,mcgovern2022we}. As with ADML, values-aware innovation in machine learning is an axis of research that is highly underappreciated.

\section*{Impact statement}

This position paper focuses on the interplay between machine learning innovation and applications with a view to real-world impact. The framework we present of stakeholder-centric design aligns with the best practices of ``AI for Good'', and our intention in this paper is to advance the paradigms of innovation that will help machine learning benefit society at large. However, it is also possible to use the application-driven ML paradigm in applications that are detrimental to society or unethical. We do not discuss at length the topic of aligning ML applications with particular goals and value systems, and point the reader instead to existing resources on this topic.

\section*{Acknowledgments}

DR, AAG, and AW were supported in part by the Canada CIFAR AI Chairs program. SB was supported in part by an MIT J-WAFS seed grant (\#2040131). DR and SB were also supported by the Global Center on AI and Biodiversity Change (NSF OISE-2330423 and NSERC 585136). ER was supported by the Harvard Data Science Initiative, the Center for Research on Computation and Society at Harvard, and Microsoft. CM was supported in part by the Environmental Data Science Innovation and Inclusion Lab (NSF DBI-2153040). We also acknowledge helpful input from Martha White, Elissa Strome, Hal Daum\'e III, and the anonymous reviewers.

\bibliography{bibliography}
\bibliographystyle{icml2024}

\end{document}